\documentclass[runningheads]{llncs}
\usepackage{amsmath}
\usepackage{graphicx}
\usepackage{comment}
\usepackage{tabularx}
\usepackage{subcaption}
\usepackage{amsmath}
\usepackage{amssymb}
\usepackage{comment}

\begin{document}
\title{A Comparison of Tiny-nerf versus Spatial 
Representations for 3d Reconstruction}
%
%
\author{Saulo Abraham Gante \orcidID{0000-0001-6012-4003} 
\and Juan Irving Vasquez \orcidID{0000-0001-8427-9333} \and Marco Antonio Valencia \orcidID{0000-0003-3990-0463}
\and Mauricio Olguín Carbajal  \orcidID{0000-0002-2296-8536}
}
\authorrunning{Gante \textit{et al}.}
%
\institute{Instituto Politécnico Nacional (IPN), Centro de Innovación y Desarrollo Tecnológico en Cómputo (CIDETEC), Ciudad de México, México.
\email{sganted1500@ipn.mx}\\}
\maketitle              
\begin{abstract}
Neural rendering has emerged as a powerful paradigm for synthesizing images, offering many benefits over classical rendering by using neural networks to reconstruct surfaces, represent shapes, and synthesize novel views, either for objects or scenes. In this neural rendering, the environment is encoded into a neural network. We believe that these new representations can be used to codify the scene for a mobile robot. Therefore, in this work, we perform a comparison between a trending neural rendering, called tiny-NeRF, and other volume representations that are commonly used as maps in robotics, such as voxel maps, point clouds, and triangular meshes. The target is to know the advantages and disadvantages of neural representations in the robotics context. The comparison is made in terms of spatial complexity and processing time to obtain a model. Experiments show that tiny-NeRF requires three times less memory space compared to other representations. In terms of processing time, tiny-NeRF takes about six times more to compute the model.

\keywords{Neural Rendering  \and NeRF \and 3D Reconstruction \and Mapping.}
\end{abstract}

\section{Introduction}

The recent and continuous advances in neural rendering have shown numerous applications and became a new field of study in the graphics community. Some of these efforts are in the implicit functions which represent shapes in three dimensions (3D) \cite{LDIF,conet}. The tool for creating the neural representations is a multi-layer perceptron (MLP). This MLP works as a general implicit function approximator. On the other hand, there are plenty of methods to reconstruct surfaces, and represent shapes and volumes in 3D space. Some examples are meshes \cite{meshes}, point clouds \cite{pointclouds}, voxel maps \cite{voxmap}, and octrees \cite{octree}. The latter offers diverse capabilities to reconstruct or create 3D models and those have diverse applications in robotics and artificial vision \cite{iv,vic,uyanik2018spgs,mur2017orb}.

In this paper, we perform a comparison between neural representations, point clouds, meshes, and voxel maps in terms of memory space and processing time required to obtain a model. The main objective is to show, clarify or put some important considerations for future works related to object reconstruction.

The MLP employed follows the architecture proposed in tiny-NeRF by the authors of Neural Radiance Fields (NeRF) \cite{nerf}. We design a grid search-based experimentation. For the tiny-NeRF,  the independent variables are i) learning rate ii) encoding functions and iii) seed;  the resulting grid search has 36 experimental units. In addition, the experiment employs the same capturing positions to create the neural model, point cloud, and voxel map. The experiment shows that neural representations require 3 times less memory to store a model but on the other hand they takes about 6 times more to compute a model concerning the other representations.

The rest of the paper is structured as follows. Section 2 introduces the required concepts. Section 3 presents the related work and advances of volume, surface and neural representations. In Section 4, we present the methodology used to perform the experiments carry out in Section 5, where results are also reported. Finally, the conclusions and the future work are given in Section 6. 

\section{Preeliminaries}

In this section, we define certain concepts required to understand the topics tackled in this paper.

A commonly used data representation is the point cloud, that is a representation of body shapes and is made by points mapped in the 3D space which are usually produced by sensors or scanners. The point cloud could be processed in order to create more accurate representations. Meshes are representations of shapes formed by a set of nodes and connections between them, one of the advantages is that it has a range of different resolutions which means that it could be as accurate as wanted but the more resolution those have the more computation it needs to complete the representation. Another 3D shape representation is the voxel, which is a cube of unitary distance, and the union of a set creates a voxel map representation. 

The voxel map represents shapes of objects or it is possible to represent the volume/solid by using a voxel carving technique. Those representations are commonly used to reconstruct shapes, objects and maps. 


According to \cite{sotanr} there is no definition for neural rendering and suggests a definition for Neural Rendering as: ``Deep image or video generation approaches that enable explicit or implicit control of scene properties such as illumination, camera parameters, pose, geometry, appearance, and semantic structure."

A recent proposed neural rendering technique is NeRF \cite{nerf}, it became one of the most popular and extensively used to render objects in 3D space due to its capabilities to create novel views in the reconstructed scene.

NeRF \cite{nerf} is an approach for creating novel view synthesis, it uses a set of input views to optimize a continuous volumetric scene function, as a result, this optimization produces a novel view of a complex scene. Its input is a 5D vector function, which contains the 3D space location(x,y,z)  and 2D viewing direction ($\theta , \phi$) and the output is an emitted color: Red, Green, Blue (r,g,b) and volume density ($\sigma$). NeRF uses the concept of encoding functions where the purpose of these functions is to map the input into a higher dimensional space where the MLP can more easily approximate higher frequency functions.

To generate a NeRF from a specific viewpoint, first, a set of rays are marched through the scene, the data generated is fed into the neural network and produce a set of RGB$\sigma$ values then the data is structured into a 2D image.

\section{Related Work}

The 3D reconstruction of spaces and objects is not a new research topic and has many approaches  which reconstruct scenes employing different techniques, the accuracy of the volumetric representations relies on the resolution employed to map, and the more resolution is wanted the more computation is needed which means more time is needed to achieve a good result.


Volume representations in 3D space have many methods to represent synthetically objects, like meshes \cite{meshes}, point clouds \cite{pointclouds}, voxel maps \cite{voxmap}, and octree \cite{octree}. Those offer diverse capabilities to reconstruct or create 3D models and those have diverse applications in robotics and artificial vision \cite{iv,vic,uyanik2018spgs,mur2017orb}. Despite the popularity that they have, resolution of the representations is one of it cons. Also, the memory consumption  between them is variable and usually requires memory in the order of Megabytes (MB).


Neural rendering has gained popularity since it employs a multi-layer perceptron (MLP) to achieve these tasks \cite{pifu,siren}. In \cite{conet,LDIF} are presented different techniques to represent shapes and volumes in 3D space. 
The proposed methods concentrate its effort in creating those representations and compare it with state-of-the-art. On the other hand, in \cite{divc} they propose an approach for volume compression and compare it with voxel maps. A Simultaneous Localization And Mapping system is proposed by \cite{sucar}, they compare it with  truncated signed distance function (TSDF) method, both approaches do a comparison in terms of memory consumption, and stand out a good performance. 

We believe that time taken in the process of the reconstructions is an important variable to take in consideration, and that the consulted approaches do not report those differences in terms of time. 

\section{Methodology}

\begin{figure}[ht]
    \centering
    \includegraphics[height = 0.3 \textheight]{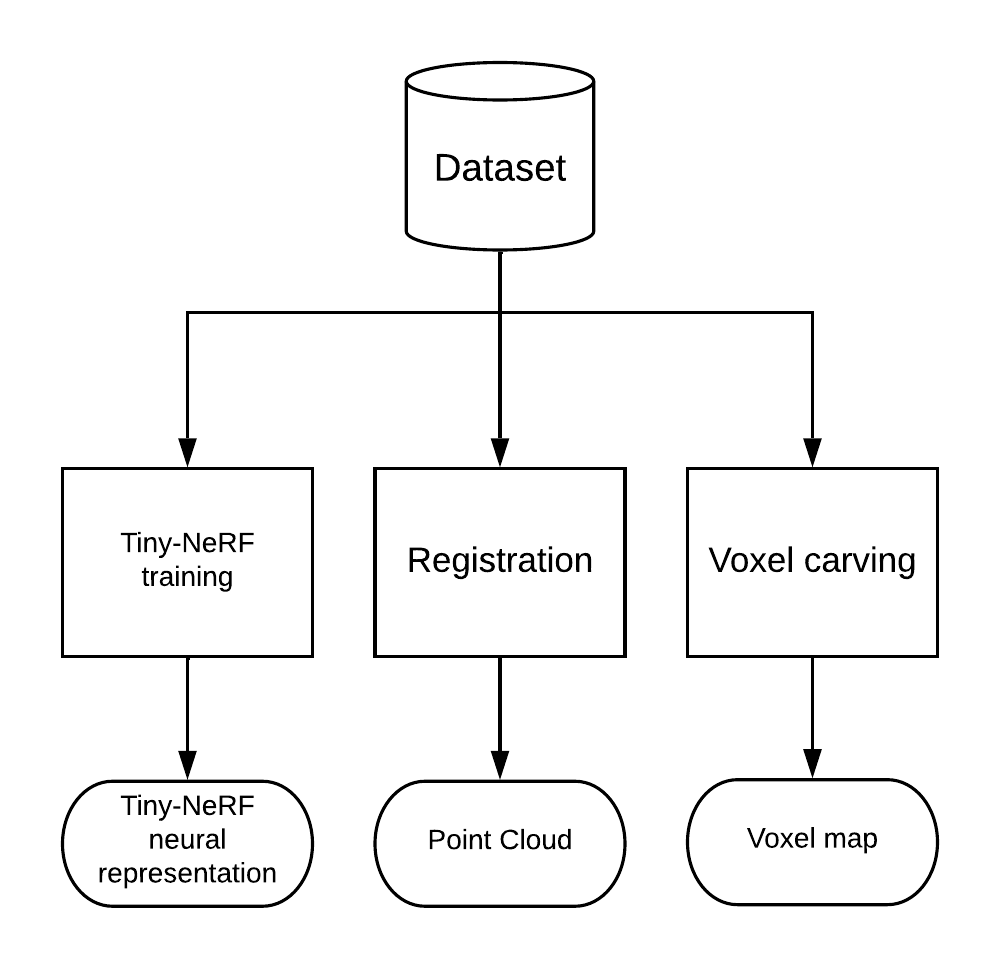}
    \caption{General diagram of the experiments. Given a dataset which contains  positions in 3D space and images, the data required is extracted with a simulator, as a result the three representations to be compared are obtained.} 
    \label{fig:diag}
\end{figure}

We want to experiment with neural representations, exploring the advantages that those have over existent representations used for 3D space reconstruction. We do a comparison between neural representations proposed by the authors of NeRF  \cite{nerf}, and three different spatial representations used to model objects, such as meshes, voxel maps, and point clouds. 
We use a dataset that contains 106 (see Fig. \ref{fig:capos}) pairs of sensor poses, and using those poses, we extract the required data in order to create the proposed representations (Figure \ref{fig:diag}). The main reason to use one dataset is that we want to give the algorithms the same point of view for a fair comparison in terms of data that could be extracted given the poses in the data set. 

\subsection{Dataset}

\begin{figure}[!h]
    \centering
    \includegraphics[height = 0.2 \textheight]{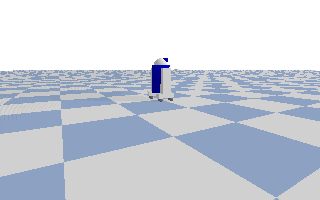}
    \caption{World setup. An object is placed in the simulated world.}
    \label{fig:simu}
\end{figure}

Given that we propose a comparison with synthetic data, we use a simulator to render images of an object (simulating a camera inside the simulated world). See Figure \ref{fig:simu}. Then, a world is needed to set up, configure it with a ground, and place the mesh of the object of interest in it. From the synthetic datasets used in NeRF \cite{nerf}, we apply transformation matrices as in Equation (\ref{eq:transm}) 

\begin{figure}[ht]
    \centering
    \includegraphics[height = 0.3 \textheight]{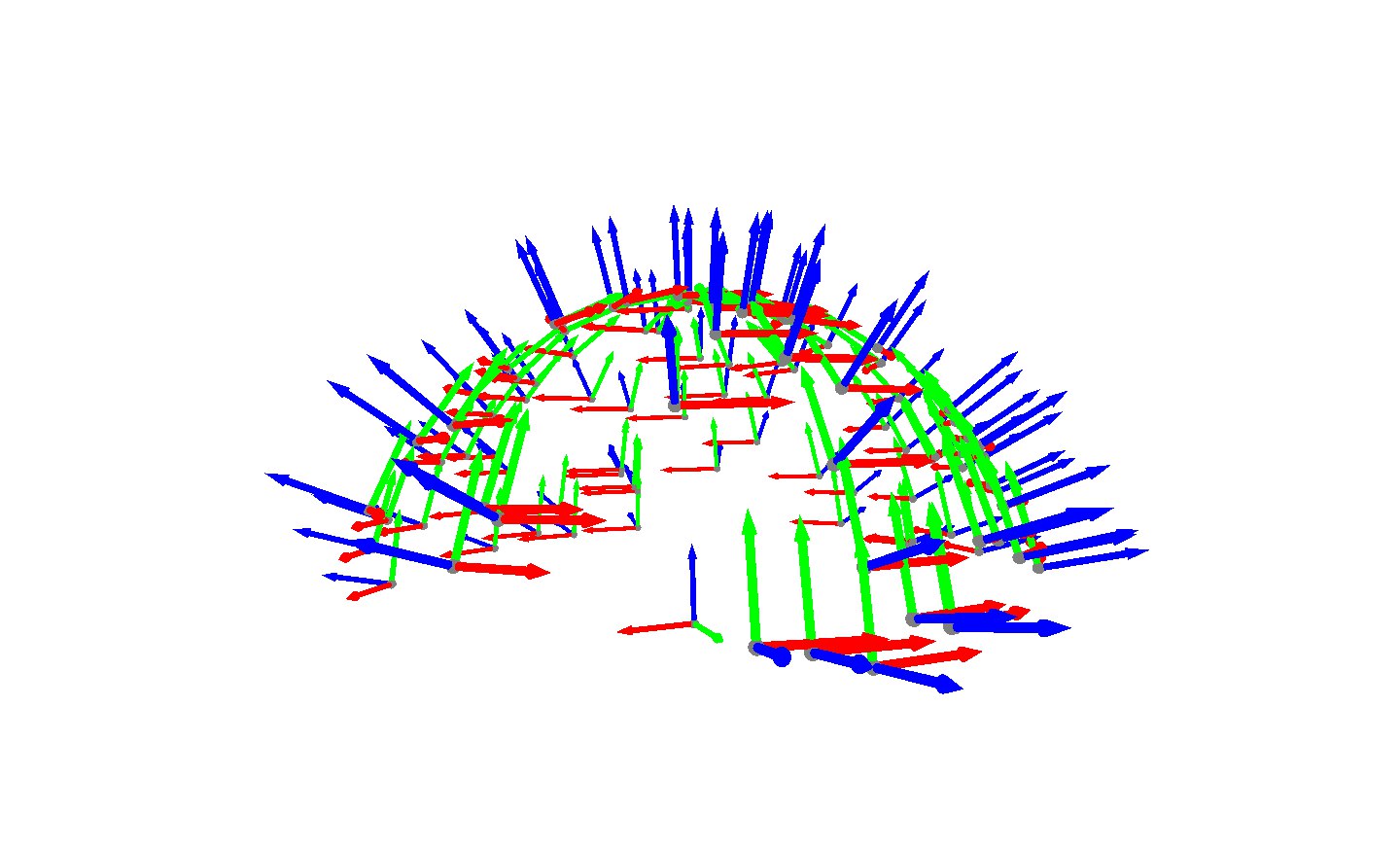}
    \caption{Capturing poses. In this figure it is shown the 106 poses used, the frame of reference is the described by OpenGL \cite{opengltuto} for synthetic cameras. }
    \label{fig:capos}
\end{figure}

\begin{equation}
T = \begin{pmatrix}
R & p \\
0 & 1
\end{pmatrix}
\label{eq:transm} ,
\end{equation}
where $R$ indicates the rotation matrix, whose values represent the rotations over the three axes, and the $p$ indicates the position vector, whose values contain the position of a body in a 3D space $(x,y,z)$. Please see Figure \ref{fig:capos}.

Having those positions in 3D space the required datasets are extracted, that is RGB images, ray casting points, and depth data. All in order to create the proposed reconstructions. 

\subsection{Point cloud and voxel map} 

Open3D library \cite{Zhou2018} allows us to visualize objects and create representations. For the point cloud, it is created by the use of ray-tracing which emits synthetic rays in simulation when those touches or intersect with a surface return a value, having this is possible to calculate in $\mathbb{R}^3$ and map them into points in space, creating a point cloud. This process is repeated every capture, then the resulting points are concatenated and filtered to reduce possible noise created by captures.

We create a voxel model using the technique of \emph{voxel carving}, using a pinhole camera and homogeneous transformation matrix is possible to create a voxel dense given the resulting images and a silhouette to employ a \emph{carve silhouette} method provided by Open3D, resulting in a voxel model.

\subsection{Tiny-NeRF}

As explained above, NeRF \cite{nerf} receives as input a set of data that express location and viewing direction where the output is an emitted color and a volume density. Tiny-NeRF is a simplified version of NeRF, which is an MLP conformed by 6 fully-connected ReLU layers each with 256 filter size, one fully-connected ReLU layer with a filter size of 64 then an output layer that expresses the emitted RGB$\sigma$ at a certain position with a four filter size layer. The process starts by getting rays according to the pose, then the returned rays become useful to map 3D points which are going to be fed into Tiny-NeRF input, the output of the model is used to compute opacities and RGB data, finally the weights are calculated and the process is repeated.

\section{Experiments}

We evaluate the Tiny-NeRF describing a grid search where certain variables are changed over the experiments. Using the same position captures we perform reconstruction with voxels and a point cloud. All the data was synthetic and obtained using Open3D.

Our experiments run in Python and the libraries employed are Pytorch \cite{NEURIPS2019_9015} for the MLP or neural representations, Pybullet \cite{coumans2021} and Open3D \cite{Zhou2018}.The hardware employed for those experiments is the CPU/GPU provided by Colab which allows us to use a Graphic card: Tesla P100-PCIE-16GB with 16GB of GPU-RAM, 25.46 GB of RAM, and 166.83 GB in Hard disc drive.

\subsection{Tiny-NeRF training}

Tiny-NeRF is a simplified version of NeRF, which is an MLP conformed by six fully-connected ReLU layers each with a 256 filter size, one fully-connected layer with a filter size of 64 then an output that expresses the RGB$\sigma$ values. The grid search proposed to vary over three variables and the values are:

\begin{itemize}
	\item \textbf{Seed}:  2057, 5680 and 7461.
	\item \textbf{Learning rate}: $5x10^{-3}$, $5x10^{-3}$ and $5x10^{-3}$
	\item \textbf{Encoding functions} : 6, 9, 10 and 12. 
\end{itemize}

For the Neural Networks (NN) training a commonly used metric is Loss since it evaluates how bad predicts on an example, the \emph{Peak Signal-to-Noise Ratio} (PSNR) is used to measure the ratio between a signal and the noise which affects the representation of this signal; in this case, the PSNR is used to measure how well the Tiny-NeRF does a representation compared to the original images. On the other hand, to measure time the unit employed is seconds (s) and to measure space in memory we utilize MB.

\begin{figure}[!ht]
    \centering
    \subfloat[Frontal point of view.]{\includegraphics[width=0.45\textwidth]{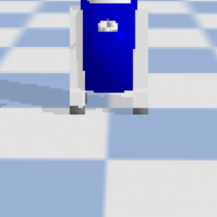}}
    \hspace{0.5cm}
    \subfloat[Lateral point of view.]{ \includegraphics[width=0.45\textwidth]{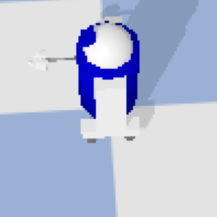}}
    \caption{Images of the object of study extracted using Pybullet\cite{coumans2021}}
    \label{fig:r2d2}
\end{figure}

To obtain the data set we employed Pybullet \cite{coumans2021} simulator which let us set simulated worlds, set objects in it (Figure \ref{fig:simu}) and create pinhole cameras to extract or create synthetic images, among other things. Once the object is set in the world, it is possible to create a synthetic camera given its position, target position, field of view (FOV), near and far plane distance, weight and height of the image. The positions are given by the data capturing positions, the FOV is $17.70^{\circ}$, the weight and height are equal to 100. Resulting in images like the ones in Figure \ref{fig:r2d2}.

The experiments with the Tiny-NeRF are iterated for five thousand epochs each and the variables are modified at each experiment, the number of experiments is 36. The results of this experiment are shown in Table \ref{table:SG}. The entire experiment took about 65,100 seconds which means that approximately every experiment took 1,808.33 seconds to be completed. To summarize the information in Table \ref{table:SG}, the average was calculated (Table \ref{table:SGA}) in order to easily extract which parameters perform better results with the MLP.

\begin{table}[!ht]
\centering
\fontsize{7}{6}
\begin{tabular}{| l | m{1.2cm}| m{2cm}| m{2cm}| m{2cm}| c | c |}
\hline
ID  & Factor 1 & Factor 2 & Factor 3 & Metric 1 & Metric 2 \\ \hline
   & Seed & Learning Rate  &  Coding functions        &   Loss     &  PSNR  (dB)        \\ \hline

1 & 2057 &$5\times10^{-3}$   &  6    & 0.5463 & 2.6257 \\ \hline
2 & 2057 &$5\times10^{-3}$   &  9    & 0.0030 & 25.2288 \\ \hline
3 & 2057 &$5\times10^{-3}$   &  10   & 0.0493 & 13.0715 \\ \hline
4 & 2057 &$5\times10^{-3}$   &  12   & 0.5463 & 2.6257 \\ \hline
5 & 2057 &$5\times10^{-4}$   &  6    & 0.5463 & 2.6257 \\ \hline
6 & 2057 &$5\times10^{-4}$   &  \textbf{9  }  & \textbf{0.0025} & \textbf{26.0206} \\ \hline
7 & 2057 &$5\times10^{-4}$   &  10   & 0.5463 & 2.6257 \\ \hline
8 & 2057 &$5\times10^{-4}$   &  12   & 0.5463 & 2.6257 \\ \hline
9 & 2057 &$5\times10^{-5}$   &  6    & 0.5463 & 2.6257 \\ \hline
10 & 2057 &$5\times10^{-5}$  &  9   & 0.0026 & 25.8503 \\ \hline
11 & 2057 &$5\times10^{-5}$  &  10  & 0.5463 & 2.6257  \\ \hline
12 & 2057 &$5\times10^{-5}$  &  12  & 0.5463 & 2.6257 \\ \hline
13 & 7461 &$5\times10^{-3}$  &  6   & 0.0921 & 10.3574 \\ \hline
14 & 7461 &$5\times10^{-3}$  &  9   & 0.0032 & 24.9485 \\ \hline
15 & 7461 &$5\times10^{-3}$  &  10  & 0.5463 & 2.6257 \\ \hline
16 & 7461 &$5\times10^{-3}$  &  12  & 0.5463 & 2.6257 \\ \hline
17 & 7461 &$5\times10^{-4}$  &  6   & 0.5463 & 2.6257 \\ \hline
18 & 7461 &$5\times10^{-4}$  &  9   & 0.0026 & 25.8503 \\ \hline
19 & 7461 &$5\times10^{-4}$  &  10  & 0.5463 & 2.6257 \\ \hline
20 & 7461 &$5\times10^{-4}$  &  12  & 0.5463 & 2.6257 \\ \hline
21 & 7461 &$5\times10^{-5}$  &  6   & 0.5463 & 2.6257 \\ \hline
22 & 7461 &$5\times10^{-5}$  &  \textbf{9}   & \textbf{0.0025} & \textbf{26.0206} \\ \hline
23 & 7461 &$5\times10^{-5}$  &  10  & 0.5463 & 2.6257 \\ \hline
24 & 7461 &$5\times10^{-5}$  &  12  & 0.5463 & 2.6257 \\ \hline
25 & 5680 &$5\times10^{-3}$  &  6   & 0.5463 & 2.6257 \\ \hline
26 & 5680 &$5\times10^{-3}$  &  9   & 0.0032 & 24.9485 \\ \hline
27 & 5680 &$5\times10^{-3}$  &  10  & 0.5463 & 2.6257 \\ \hline
28 & 5680 &$5\times10^{-3}$  &  12  & 0.0033 & 24.8149 \\ \hline
29 & 5680 &$5\times10^{-4}$  &  6   & 0.5463 & 2.6257 \\ \hline
30 & 5680 &$5\times10^{-4}$  &  9   & 0.0027 & 25.6864 \\ \hline
31 & 5680 &$5\times10^{-4}$  &  10  & 0.5463 & 2.6257 \\ \hline
32 & 5680 &$5\times10^{-4}$  &  \textbf{12 } & \textbf{0.0024} & \textbf{26.1979} \\ \hline
33 & 5680 &$5\times10^{-5}$  &  6   & 0.5463 & 2.6257 \\ \hline
34 & 5680 &$5\times10^{-5}$  &  9   & 0.0027 & 25.6864 \\ \hline
35 & 5680 &$5\times10^{-5}$  &  10  & 0.5463 & 2.6257 \\ \hline
36 & 5680 &$5\times10^{-5}$  &  12  & 0.0027 & 25.6864 \\ \hline
\end{tabular}
\caption{Grid search. ID express an identification number, the variable values employed for each experiment with Tiny-NeRF and the results expressed in terms of Loss and PSNR. }
\label{table:SG}
\end{table}

\begin{table}[!ht]
\centering
\begin{tabular}{|c|c|c|c|}
\hline
 LR & Functions & Loss & PSNR (dB) \\ \hline
 $5\times10^{-3}$ & 6  & 0.3949 & 4.0351\\  
 $5\times10^{-3}$ & 9  & 0.0031 & 25.0863\\ 
 $5\times10^{-3}$ & 10 & 0.3806 & 4.1953\\ 
 $5\times10^{-3}$ & 12 & 0.3653 & 4.3735\\ 
 $5\times10^{-4}$ & 6  & 0.5463 & 2.6256\\ 
 \textbf{ $5\times10^{-4}$ }& \textbf{9 } & \textbf{0.0026 }& \textbf{25.8502}\\ 
 $5\times10^{-4}$ & 10 & 0.5463 & 2.6256\\ 
 $5\times10^{-4}$ & 12 & 0.365  & 4.3770\\ 
 $5\times10^{-5}$ & 6  & 0.5463 & 2.6256\\ 
 \textbf{ $5\times10^{-5}$ } & \textbf{9}  & \textbf{0.0026} & \textbf{25.8502}\\ 
 $5\times10^{-5}$ & 10 & 0.5463 & 2.6256\\ 
 $5\times10^{-5}$ & 12 & 0.3651 & 4.3758\\ \hline
     
\end{tabular}
\caption{Average of the results in grid search.}
\label{table:SGA}
\end{table}

\begin{figure}[!ht]
    \centering
    \subfloat[Frontal point of view.]{\includegraphics[width=0.45\textwidth]{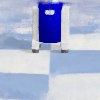}} 
    \hspace{0.5cm}
    \subfloat[Lateral point of view.]{ \includegraphics[width=0.45\textwidth]{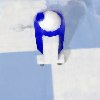}}
    \caption{Neural representations created with the Tiny-NeRF.}
    \label{fig:r2d2imp}
\end{figure}

Analysis of experiments showed that nine coding functions help the Tiny-NeRF to accurately (Figure \ref{fig:r2d2imp}) create a neural representation of the object, and the learning rate helped to achieve good performance in fewer epochs. Additionally, the neural representations took \textbf{1,808.33 seconds} to complete an experiment, and the memory space to store a representation is \textbf{1.5 MB}.

\begin{figure}[!ht]
    \centering
    \subfloat[Simulation captures.]{\includegraphics[width=\textwidth]{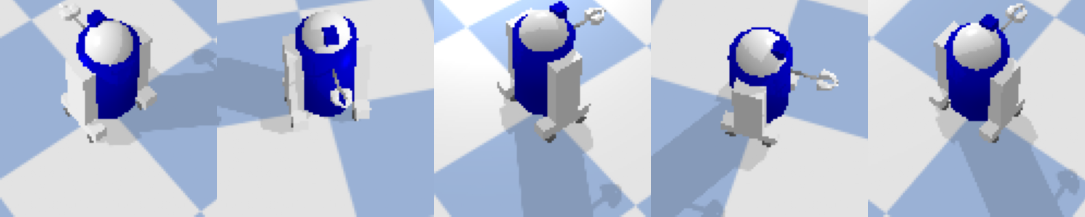}} \\
    \subfloat[Neural representations.]{ \includegraphics[width=\textwidth]{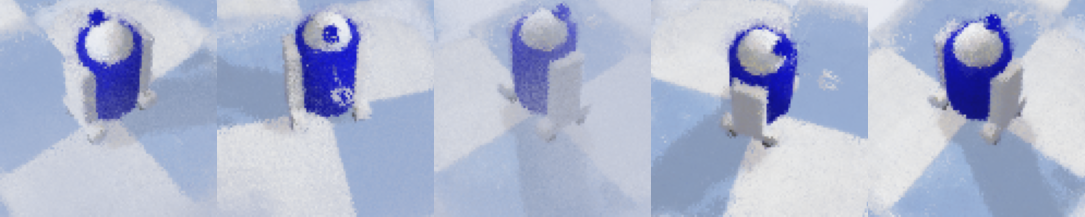}}
    \caption{Qualitative results. We extracted different posses and visualizations using Tiny-NeRF.}
    \label{fig:impvorg}
\end{figure}

Additionally to PSNR, we perform evaluations over two more metrics SSIM  and LPIPS \cite{ssim,lpips} which are commonly used to measure distances over images, looking for a measure of how well the Tiny-NeRF is rendering views. Comparing images like the ones in Fig. \ref{fig:impvorg} the metrics proposed gave as a result \textbf{0.8481} and \textbf{0.0565}, respectively. Those results affirm that the representations are good in quality but it could improved. 

\subsection{Comparison of Tiny-NeRF versus spatial representations}

Once Tiny-NeRF has been trained and the representations were created, we compared the time taken to do a representation. To measure the time, it was printed every time a process started and finished the difference between those shows the time taken. The space in memory is measured by the file space in memory that is required to store the representations.


The data employed to reconstruct was obtained by capturing in the positions of the data set, mentioned above, once the captures are done the process of data was done employing Open3D \cite{Zhou2018}.

\begin{figure}[!ht]
    \centering
    \subfloat[Point cloud reconstruction.]{ \includegraphics[width=0.45\textwidth]{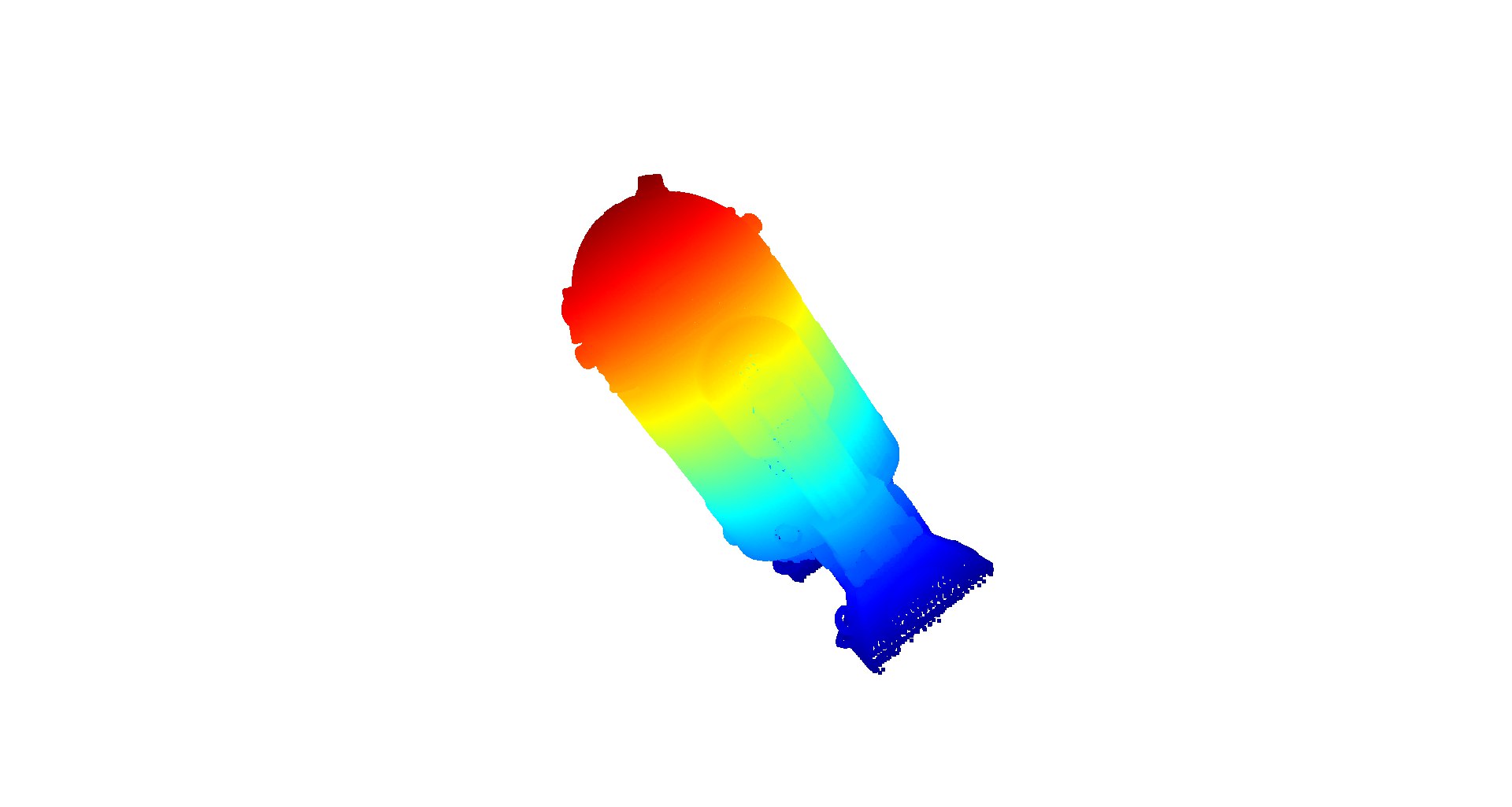}} \hspace{0.5cm}
    \subfloat[Voxel reconstruction.]{ \includegraphics[width=0.45\textwidth]{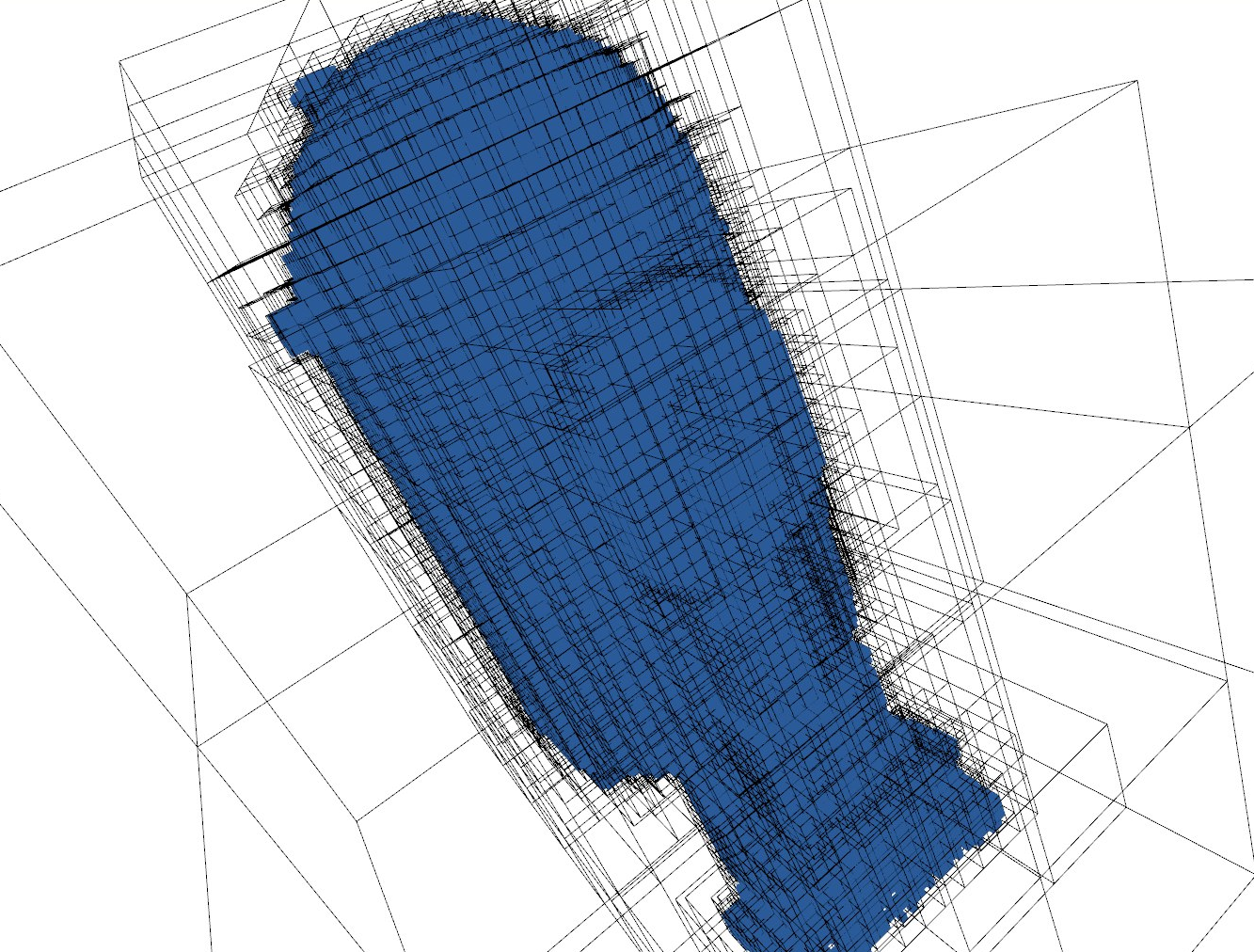}}
    \caption{Volumetric reconstructions.}
    \label{fig:nubvox}
\end{figure}

The point cloud (Figure \ref{fig:nubvox} (a)) was obtained by mapping the points resultant of a ray-tracing operation into XYZ or 3D space, those points are concatenated and finally filtrated to avoid noise in the reconstruction. The experiment took about \textbf{2 seconds} and the memory space needed is \textbf{12 MB}.

The resultant voxel map (Figure \ref{fig:nubvox} (b)) was created with the voxel carving method which not only reconstructs the surface of an object, it creates a voxel map that is a cube of certain dimensions and according to the visualized data, the algorithm carves the shape into it, creating a solid voxel representation. The experiment took about \textbf{166 seconds} and the memory space needed is \textbf{21.2 MB}.

The performed experiments results showed that implicit or neural representation requires at least 3 times less memory compared with other representations (Table \ref{tab:comp_tam}). In terms of time to process a representation, point clouds and voxel maps build the representations in about 6 times less time than the implicit representations (Table \ref{tab:comp_tim}).

\begin{table}[tb]
\center
\begin{tabular}{lc}
Representation   & Size (MB)  \\\hline
Meshes 				     &  4.5  \\
Point cloud 		     &  12.0  \\
Voxelization		     &   21.2 \\
\textbf{Implicit representation} &  \textbf{ 1.5} \\\hline
\end{tabular}
\caption{Comparative of memory size.}
\label{tab:comp_tam}
\end{table}

\begin{table}[tb]
\center
\begin{tabular}{lc}
Representation  & Time (s)  \\\hline
Meshes 				     &  28800  \\
\textbf{Point cloud} 	 &  \textbf{2} \\
Voxelization 		     &  166  \\
Implicit representation  &   1008 \\\hline
\end{tabular}
\caption{Comparative of time taken to perform a representation.}
\label{tab:comp_tim}
\end{table}

\section{Conclusions and Future work}
This paper tackles the trend research topic, neural rendering, which has many advances in graphics generation. We compare a simplified version of NeRF with different volume representations commonly used in robotics and vision reconstruction tasks, all compared in terms of memory space and time to build representations. First, we experimented with Tiny-NeRF that computes the colors over a certain position with a viewing direction; the experiments were conducted by a grid search looking to perform good representations of an object. In addition, we perform a reconstruction using voxels and point clouds. The comparison, in terms of memory space and time, shows that the Tiny-NeRF architecture (MLP) requires less memory but takes more time to build a representation. On the other hand, this experiment showed that the neural representation relies on the fine-tuning of the variables implied in the training of the MLP. We believe that the results of the experiments can offer some relevant information or considerations to take when a reconstruction task is needed. In a future work we will experiment with more objects and with a mobile manipulator robot.

%
%
%

\bibliographystyle{splncs04}
\bibliography{biblio}

\end{document}